\pdfoutput=1

\documentclass[11pt]{article}

\usepackage[final]{acl}
\usepackage{times}
\usepackage{latexsym}
\usepackage[T1]{fontenc}
\usepackage[utf8]{inputenc}
\usepackage{microtype}
\usepackage{inconsolata}
\usepackage{graphicx}
\usepackage{hyperref}
\PassOptionsToPackage{table}{xcolor}
\usepackage[table]{xcolor}
\usepackage{url}
\usepackage{multirow}
\usepackage{graphicx}
\usepackage{listings}
\usepackage{algorithm}
\usepackage{tcolorbox}
\tcbuselibrary{documentation}
\tcbuselibrary{listingsutf8}
\usepackage{booktabs}
\usepackage{pifont}
\usepackage{amsmath}
\usepackage{ascii}
\usepackage[utf8]{inputenc}
\usepackage{pmboxdraw}
\usepackage{newunicodechar}
\usepackage{subcaption} 
\usepackage{colortbl}

\usepackage{algorithmic}
\usepackage{chngcntr} 
\usepackage{amssymb}

\newcommand{\Ours}{\textsc{SetR}}
\newcommand{\ours}{\textsc{Set selection}}

\title{Shifting from Ranking to Set Selection for Retrieval Augmented Generation}

\author{
Dahyun Lee\stepcounter{footnote}\thanks{These authors contributed equally to this work.}\textsuperscript{1}
\qquad Yongrae Jo\footnotemark[2]\textsuperscript{1}
\qquad Haeju Park\footnotemark[2]\textsuperscript{1}
\qquad Moontae Lee\textsuperscript{1,2}
\\
\textsuperscript{1}LG AI Research,
\textsuperscript{2}University of Illinois Chicago
\\
 \small{
   \texttt{\href{leedhn@lgresearch.ai}{\{leedhn}},~\texttt{\href{yongrae.jo@lgresearch.ai}{yongrae.jo}},~\texttt{\href{haeju.park@lgresearch.ai}{haeju.park}},~\texttt{\href{moontae.lee@lgresearch.ai}{moontae.lee\}@lgresearch.ai}}
 }
}

\begin{document}
\maketitle
\begin{abstract}

Retrieval in Retrieval-Augmented Generation (RAG) must ensure that retrieved passages are not only individually relevant but also collectively form a comprehensive set.
Existing approaches primarily rerank top-$k$ passages based on their individual relevance, often failing to meet the information needs of complex queries in multi-hop question answering.
In this work, we propose a set-wise passage selection approach and introduce \Ours, which explicitly identifies the information requirements of a query through Chain-of-Thought reasoning and selects an optimal set of passages that collectively satisfy those requirements.
Experiments on multi-hop RAG benchmarks show that {\Ours} outperforms both proprietary LLM-based rerankers and open-source baselines in terms of answer correctness and retrieval quality, providing an effective and efficient alternative to traditional rerankers in RAG systems.
The code is available at~\url{https://github.com/LGAI-Research/SetR} 

\end{abstract}

\section{Introduction}

Bridging parametric knowledge with external information is vital for ensuring accurate and reliable generation in language models.
Retrieval-Augmented Generation (RAG) overcomes critical limitations of Large Language Models (LLMs), particularly their inability to incorporate up-to-date or domain-specific knowledge without retraining~\cite{mallen-etal-2023-trust}.
The risk of generating hallucinations is another concern when the model capacity is not sufficiently large~\cite{Huang_2024}.
By integrating an external retrieval system that provides contextually relevant and grounded evidence in real time, RAG improves both the accuracy and reliability of knowledge-intensive tasks.


A critical component of RAG systems is the retrieval and reranking module, as the quality of the retrieved information directly influences the accuracy and relevance of the generated answers~\cite{shi2023largelanguagemodelseasily,wu2024pandorasboxaladdinslamp,wu2024clashevalquantifyingtugofwarllms,wadhwa2024ragsrichparametersprobing,hong2024gullibleenhancingrobustnessretrievalaugmented,feng2024donthallucinateabstainidentifying}.
Numerous studies have explored optimizing the effective integration of LLMs and retrieval modules. ~\citet{asai2023selfraglearningretrievegenerate,jeong2024adaptiveraglearningadaptretrievalaugmented} focus on dynamically determining the necessity of retrieval and when to stop it. 
~\citet{wang2023shallpretrainautoregressivelanguage,shao2023enhancingretrievalaugmentedlargelanguage} involve alternating between retrieval and generation, enriching contextual references through multiple retrieval iterations. 
~\citet{trivedi2023interleavingretrievalchainofthoughtreasoning,sarthi2024raptorrecursiveabstractiveprocessing} iteratively decompose and refine complex questions, addressing them through retrieval and generation.
However, these multi-step approaches require significantly more resources, potentially limiting their feasibility for real-world applications.
Therefore, many existing RAG systems utilize conventional reranking modules with a straightforward top-k selection strategy, originally developed and optimized for search applications.

\begin{figure*}[ht]
\includegraphics[width=\linewidth]{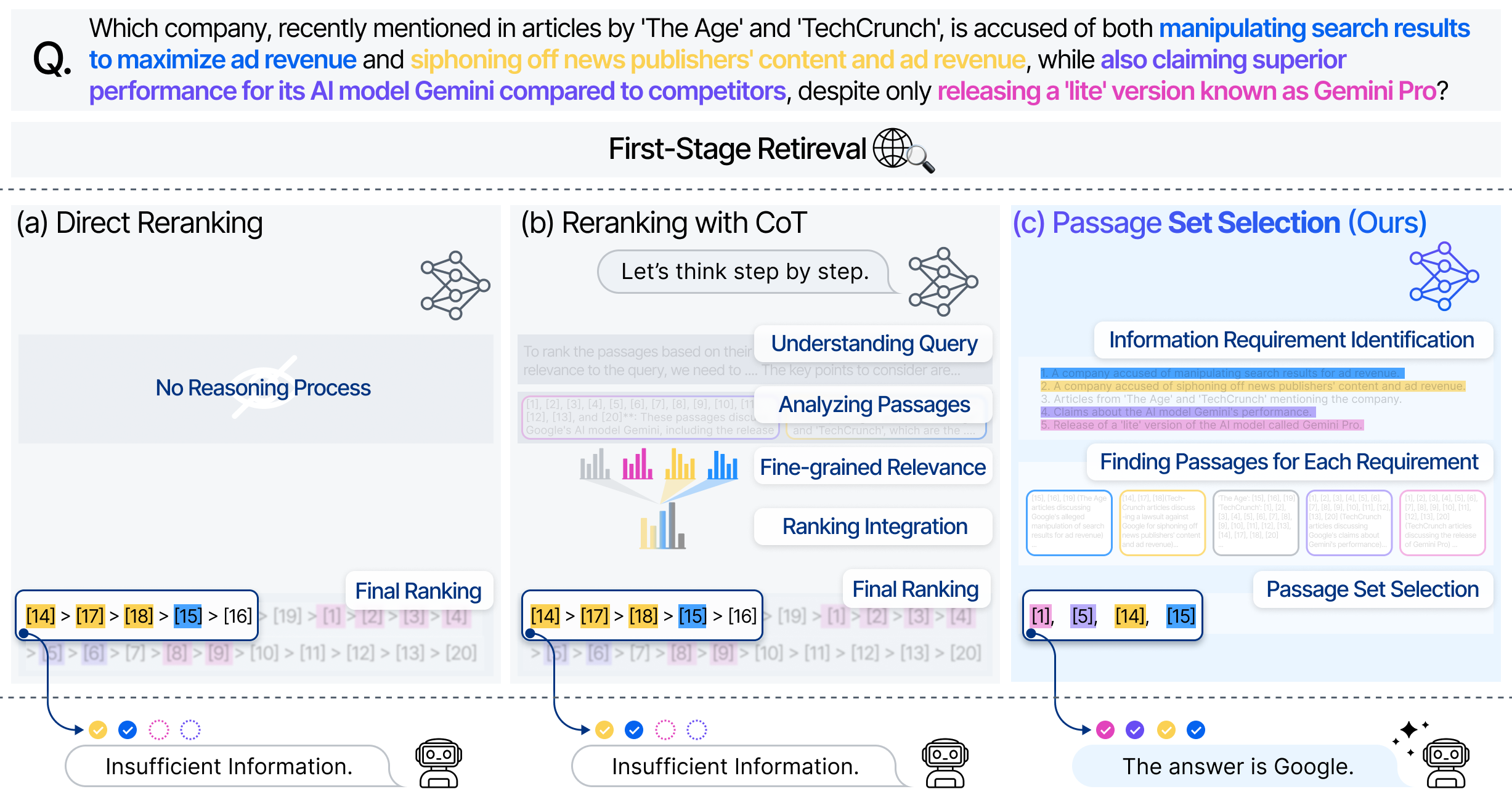}
\caption{Overview of our~\ours~approach for RAG system, compared to passage reranking methods:
\textbf{(a) Direct Reranking}, \textbf{(b) Reranking with CoT}, and \textbf{(c) Our \ours}.
(a) lacks explicit reasoning, making it unclear whether multiple aspects are considered.
(b) incorporates a reasoning process but may misrepresent or underemphasize key information when integrating relevance for final ranking.
(c) \ours~explicitly identifies all necessary information and selects relevant passages, ensuring more comprehensive information.
This example is from the MultiHopRAG dataset~\cite{tang2024multihoprag}.}
\label{main_figure}
\end{figure*}

We argue that RAG systems have distinct information demands that set them apart from traditional search engines.
While traditional search engines rank individual results by relevance, RAG systems need a curated set of passages to generate accurate answers, requiring not only relevance but also diversity, completeness, and the comprehensiveness of retrieved passages.
For example, when identifying a company based on a mix of business strategies, controversies and product claims, a RAG system must retrieve a diverse range of sources covering all these aspects.
If the system retrieves only passages discussing product claims but not those covering controversies, business strategies, and other relevant factors, it may produce an incomplete or inaccurate response.

To address these challenges, we propose a set-wise passage selection approach that optimizes the quality of the passage set as a whole, rather than treating retrieval as an independent ranking task. 
This holistic strategy encourages comprehensive coverage of essential information while reducing redundancy within the selected set (Figure~\ref{main_figure}).

To make it practical for real-world applications, we introduce {\Ours}, a fine-tuned LLM designed to implement our set-wise passage selection approach.
The model first analyzes the user question using Chain-of-Thought (CoT) reasoning to identify its information requirements, and selects an optimal subset from the full list of retrieved passages, maximizing coverage and relevance. This enables our method to serve as an effective alternative to rerankers in RAG systems.

Experiments on multi-hop RAG benchmarks, including HotpotQA~\citep{yang2018hotpotqadatasetdiverseexplainable}, 2WikiMultiHopQA~\citep{xanh2020_2wikimultihop}, MusiQue~\citep{trivedi2021musique}, and MultiHopRAG~\citep{tang2024multihoprag}, demonstrate that set-wise passage selection significantly enhances the effectiveness of RAG systems. It outperforms both proprietary LLM-based reranking and open-source rerankers, achieving higher answer correctness. 
Moreover, retrieval performance evaluation on MultiHopRAG~\citep{tang2024multihoprag} indicates improvements in precision and recall, further underscoring its strong retrieval capabilities even in isolation.

The ablation study of {\Ours} reveals that the performance boost stems from both the set-wise passage selection approach and CoT reasoning for identifying information requirements, each making a distinct and effective contribution to retrieval quality. 
The analysis shows that both components enhance information coverage in the retrieved set while effectively rejecting negative candidates.

The contributions of this work are threefold:
\begin{itemize}  
    \item \textbf{Set-wise Passage Selection for RAG:} We propose an information requirement-based set-wise passage selection approach that ensures collective coverage of retrieved passages, optimizing the retrieved set as a whole rather than treating retrieval as an independent ranking task.

    \item \textbf{Comprehensive Evaluation of Set Retrieval and Generation:} To validate the effectiveness of our approach, we conduct extensive evaluations on both the retrieved passage sets and the final generated outputs. Our experiments on multi-hop RAG benchmarks demonstrate that our method outperforms both proprietary LLM-based rerankers and open-source alternatives, achieving better answer correctness, while also improving retrieval precision and recall.

    \item \textbf{Open-Source Contribution:} We release the complete and fully reproducible recipe of {\Ours}, implementing our set-wise passage selection approach. We hope this work facilitates future research and communty-driven advancements in retrieval strategies for RAG systems.
\end{itemize}










\section{Related Work}

\subsection{Retrieval-Augmented Generation (RAG)}
Retrieval-Augmented Generation (RAG) systems combine retrieval modules with language models to enhance factual accuracy and reduce hallucinations~\cite{lewis2021retrievalaugmentedgenerationknowledgeintensivenlp, guu2020retrieval}.
Standard pipelines typically employ a first-stage retriever such as BM25~\cite{10.1561/1500000019} or DPR~\cite{karpukhin2020dense} followed by a reranking module that estimates relevance via pointwise, pairwise, or listwise strategies~\cite{nogueira2019passage,qin2023large,zhuang2023rankt5, yoon2024listt5}.

Recently, Large Language Models (LLMs) have been applied to listwise reranking in RAG systems through prompting and distillation~\cite{sun2024chatgptgoodsearchinvestigating,pradeep2023rankzephyreffectiverobustzeroshot}, demonstrating strong performance. 
However, these models primarily focus on individual passage relevance and often overlook set-level properties such as diversity or coverage, which are crucial for generating complete and accurate answers in multi-hop or compositional question answering tasks~\cite{tang2024multihoprag}.

\subsection{Refinement and Iteration in RAG}
To address the limitations of ranking-based retrieval, recent studies have explored retrieval strategies better aligned with the needs of RAG.
These include adaptive retrieval based on query complexity~\cite{jeong2024adaptiveraglearningadaptretrievalaugmented}, multi-step reasoning with agent-based ranking~\cite{niu2024judgerank}, and context pruning~\cite{chirkova2025provence}.
Iterative methods such as Self-RAG~\cite{asai2023selfraglearningretrievegenerate} and CoRAG~\cite{wang2024coragcostconstrainedretrievaloptimization} refine queries or perform multi-round retrieval to improve relevance.

While effective, these approaches often entail substantial computational overhead and are sensitive to prompt design and hyperparameter choices~\cite{asai2023selfraglearningretrievegenerate, niu2024judgerank}.
In contrast, our study provides a lightweight and efficient alternative by selecting a coherent subset of passages in a single step through explicit modeling of information requirements.
This set-oriented approach is compatible with iterative retrieval pipelines and can also function as a direct replacement for conventional rerankers in standard RAG systems.

\section{\Ours: Set-wise Passage Selection for Retrieval-Augmented Generation}
In this section, we present \Ours~(\underline{Set}-wise passage selection for \underline{R}etrieval-Augmented Generation), a novel retrieval paradigm that moves beyond conventional reranking strategies.
We begin by formalizing the passage selection task and motivating the need for a set-oriented perspective~(§\ref{subsec:task_definition}).
Next, we present our information requirement identification (IRI) method, which utilizes Chain-of-Thought (CoT) reasoning to guide passage selection~(§\ref{subsec:iri}).
Finally, we describe the architecture and training methodology of~\Ours, a distilled model fine-tuned for efficient set-wise passage selection~(§\ref{subsec:ours}), along with the data construction details (§\ref{subsec:data}) and training procedures (§\ref{subsec:training}). 

\subsection{Task Definition}
\label{subsec:task_definition}



We define the passage retrieval task as the process of selecting an optimal set of passages from a pool of retrieved candidate passages to address a specific information need, such as supporting precise and coherent responses in RAG systems. Traditionally, this task has been framed as a reranking problem, where passages are scored individually and the top-$k$ results are selected based on their relevance scores.

However, we argue that relevance-based reranking alone is insufficient for retrieval modules in RAG systems, which require more holistic retrieval strategies. 
To address this, we propose a set-wise retrieval approach that jointly optimizes the relevance, completeness, and conciseness of the retrieved set. This method also eliminates the need to manually select the top-$k$ value in reranking, streamlining the process.

\subsection{Information Requirement Identification via CoT Reasoning}
\label{subsec:iri}

We design a prompting strategy that enables set-wise passage selection by systematically identifying information requirements through a structured, step-by-step reasoning process. 
As illustrated in Figure~\ref{fig:extraction}, our prompt guides a zero-shot CoT reasoning process that decomposes the input question into distinct information subgoals.
The key process consists of three key steps: 
(1) enumerating the key information requirements necessary to answer the question;
(2) identifying passages that contain relevant information for each requirement; and 
(3) selecting a subset of passages that collectively provide the most comprehensive and diverse coverage to effectively answer the query. 

By prompting Large Language Models (LLMs) with both the question and a set of candidate passages retrieved in an earlier retrieval stage, this method enables fine-grained analysis and effective set selection.
In contrast to zero-shot listwise reranking, our approach imposes no constraint on the inclusion of all candidate passages and does not enforce any ranking or ordering in the final selection.

\subsection{Model Distillation}
\label{subsec:ours}

To ensure the proposed approach is practical for real-world applications, we train {\Ours}, a distilled model fine-tuned for the set-wise passage selection task through information requirement identification. While proprietary LLMs exhibit strong performance, their cost and latency would make their use in real-time search systems impractical. 
Instead, we distill step-by-step reasoning ability into a specialized, lightweight model for efficiency.

\subsubsection{Data Construction}
\label{subsec:data}
For distillation, we construct a dataset based on 40K training questions\footnote{\url{https://huggingface.co/datasets/castorini/rank_zephyr_training_data}} from \citet{pradeep2023rankzephyreffectiverobustzeroshot}, originally derived from the MS MARCO v1 passage ranking dataset~\citep{sun2024chatgptgoodsearchinvestigating}. 
Each query is paired with the top-20 retrieved candidate passages. 
We then apply set-wise passage selection to generate teacher-labeled selections, which are subsequently distilled into our student model. 
To perform this labeling, we use GPT-4o with a zero-shot prompting approach.
Following \citet{pradeep2023rankvicunazeroshotlistwisedocument}, we replaced all instances of [n] in passages with (n) to prevent model confusion during data synthesis and inference. 
We used the fix\_text function from ftfy\footnote{\url{https://pypi.org/project/ftfy}} to preprocess all inputs before feeding them into the model.

\subsubsection{Training}
\label{subsec:training}
We adopt Llama-3.1-8B-Instruct\footref{llama3.1-8b-inst} as the base model and train it using a standard supervised fine-tuning approach. 
The input consists of a prompt including the user question and retrieved passages, while the output includes the CoT reasoning from the teacher model along with the selected passages.

For the ablation study, we present two additional model variations. We refer to the original model as {\Ours}-CoT \& IRI for comparison with these variations. Full prompt details are provided in Appendix~\ref{sec:appendix_implementation_details}.
\begin{itemize}  
\item \textbf{{\Ours}-Selection only} is a model trained to generate only the final selected passages without any reasoning process. 
\item \textbf{{\Ours}-CoT} is a model trained with general CoT reasoning using a standard ``Let's think step-by-step prompt'' prompt, but does not explicitly identifying distinct information requirements. 
\item \textbf{{\Ours}-CoT \& IRI} is the full model that incorporates both CoT reasoning and explicit information requirement identification, and performs passage selection accordingly.
\end{itemize}

\begin{figure}
\begin{tcolorbox}[
    colback=gray!10,
    colframe=black,
    fontupper=\ttfamily\footnotesize,
    title={Passage Selection Prompt of {\Ours}},
]

I will provide you with \{num\} passages, each indicated by a numerical identifier []. Select the passages based on their relevance to the search query: \{question\}.\\
\\
\{context\}
\\
\\
Search Query: \{question\}
\\
\\
Please follow the steps below:\\
Step 1. Please list up the information requirements to answer the query.\\
Step 2. for each requirement in Step 1, find the passages that has the information of the requirement.\\
Step 3. Choose the passages that mostly covers clear and diverse information to answer the query. Number of passages is unlimited. The format of final output should be `\#\#\# Final Selection: [] []', e.g., \#\#\# Final Selection: [2] [1].
\end{tcolorbox}
\vspace{-1em}
\caption{The set-wise passage selection prompt with Chain-of-Thought information requirement identification process for \Ours.}
\label{fig:extraction}
\end{figure}

\begin{table*}[ht]\footnotesize
\centering
\resizebox{\linewidth}{!}{
\begin{tabular}{ll|c|cccccccc}
\toprule
    \multirow{2}{*}{\textbf{Retrieval}} &
    \multirow{2}{*}{\textbf{Model}} & \textbf{\# of } & \multicolumn{2}{c}{\textbf{HotpotQA}} & \multicolumn{2}{c}{\textbf{2WikiMultiHopQA}} & \multicolumn{2}{c}{\textbf{MuSiQue}} & \multicolumn{1}{c}{\textbf{MultiHopRAG}} \\
    & & \textbf{Passages} &EM&F1&EM &F1&EM&F1&Accuracy \\
\noalign{\hrule height 0.8pt}

\multirow{12}{*}{BM25} 
& \multicolumn{9}{l}{\cellcolor{lightgray!40}\textsc{Retrieval only}} \\

& - &  5.00 & 26.90 & 25.86 & 29.79 & 21.79 & 5.46 & 8.22 & 39.20 \\
& \multicolumn{9}{l}{\cellcolor{lightgray!40}\textsc{Reranking}} \\

& bge-reranker-large  & 5.00 & 29.71 & 28.08 & 30.16 & 21.84 & 6.12 & 10.00 & 42.13 \\
& RankLlama & 5.00 & 29.48 & 27.82 & 30.30 & 21.91 & 6.04 & 9.26 & 42.09 \\
& RankVicuna  & 5.00 & 28.69 & 27.31 & 30.46 & 22.42 & 5.99 & 9.03 & 40.53 \\
& RankZephyr  & 5.00 & 28.96 & 27.76 & 30.29 & 22.34 & 6.78 & 10.03 & 40.10 \\
& FirstMistral  & 5.00 & 26.71 & 26.10 & 30.15 & 21.97 & 5.29 & 8.42 & 40.29 \\
& RankGPT (\texttt{gpt-4o})  & 5.00 & 30.89 & 29.24 & 31.71 & 23.31 & 6.91 & 9.98 & \textbf{44.36} \\

& \multicolumn{9}{l}{\cellcolor{lightgray!40}\textsc{Set Selection (Ours)}} \\

& SetR-Selection only & 2.95 & \underline{31.61} & \underline{30.55} & \textbf{32.22} & 24.20 & \textbf{8.02} & \textbf{11.07} & 43.62 \\
& SetR-CoT  & 2.48 & 30.79 & 30.12 & 32.07 & \textbf{24.43} & \underline{7.03} & \underline{10.87} & 41.63 \\
& SetR-CoT \& IRI & 2.63 & \textbf{32.20} & \textbf{30.57} & \underline{32.17} & \underline{24.22} & 6.62 & 10.57 & \underline{44.13} \\
\hline
\noalign{\hrule height 0.8pt}


\multirow{12}{*}{bge-large-en-v1.5}  
& \multicolumn{9}{l}{\cellcolor{lightgray!40}\textsc{Retrieval only}} \\
& -  & 5.00 & 30.07 & 30.97 & 31.17 & 25.22 & 7.44 & 10.78 & 41.82 \\
& \multicolumn{9}{l}{\cellcolor{lightgray!40}\textsc{Reranking}} \\
& bge-reranker-large  & 5.00 & 32.48 & 33.24 & 31.92 & 25.47 & 8.06 & 12.50 & 43.50 \\
& RankLlama & 5.00 & 31.88 & 32.95 & 32.24 & 25.78 & 7.61 & 11.77 & 43.51 \\
& RankVicuna& 5.00 & 32.08 & 32.83 & 32.66 & 26.85 & 7.78 & 11.35 & 42.76 \\
& RankZephyr  & 5.00 & 31.83 & 32.97 & 32.68 & 26.59 & 8.02 & 11.72 & 41.55 \\
& FirstMistral  & 5.00 & 30.10 & 31.07 & 31.43 & 25.31 & 6.53 & 10.64 & 42.05 \\
& RankGPT (\texttt{gpt-4o}) & 5.00 & 33.85 & 34.45 & 34.36 & 28.06 & 9.43 & 13.25 & 45.69 \\
& \multicolumn{9}{l}{\cellcolor{lightgray!40}\textsc{Set Selection (Ours) }} \\
& SetR-Selection only & 3.41 & \textbf{36.68} & 37.84 & 34.84 & 29.40 & \underline{10.38} & \underline{15.28} & \underline{46.20} \\
& SetR-CoT & 2.88 & 36.46 & \textbf{38.20} & \underline{35.34} & \underline{30.34} & 9.76 & 14.31 & 45.26 \\
& SetR-CoT \& IRI  & 2.91 & \underline{36.62} & \underline{38.11} & \textbf{35.44} & \textbf{30.35} & \textbf{10.79} & \textbf{15.43} & \textbf{47.14} \\
\bottomrule
\end{tabular}
}
\caption{End-to-end question answering results across various ranking models. Each model applies reranking or selection over the top-20 passages retrieved using either BM25 or bge-large-en-v1.5. The \textbf{bold} and \underline{underlined} indicate the best and second-best performances respectively. "\# of Passages" indicates the average number of passages included in the prompt context during answer generation.}
\label{tb:RAG}
\end{table*}

\section{Experiments}
In this section, we first introduce the experimental setup (§\ref{sec:setup}). Then, we show the results for both the generation (§\ref{sec:qa_eval}) and retrieval stages (§\ref{sec:retrieval_eval}), highlighting the effectiveness of the proposed {\Ours}. More details are shown in Appendix~\ref{sec:appendix}.
\subsection{Setup}
\label{sec:setup}
\noindent\textbf{Benchmarks}. 
For evaluation, we conduct experiments in two folds: (1) end-to-end QA, and (2) retrieval task. 
For comprehensive evaluation, we adopt four widely used complex multi-hop QA datasets: HotpotQA~\citep{yang2018hotpotqadatasetdiverseexplainable}, 2WikiMultiHopQA~\cite{xanh2020_2wikimultihop}, MuSiQue~\cite{trivedi2021musique}, and MultiHopRAG~\cite{tang2024multihoprag}. 
These datasets cover diverse question types and multi-hop reasoning scenarios, offering a comprehensive evaluation of QA models in complex, real-world contexts.

\noindent\textbf{Baselines}.
We compare {\Ours}~with state-of-the-art ranking baselines across different model categories. Specifically, we include traditional unsupervised ranking models such as BM25~\cite{10.1561/1500000019}, supervised dense ranking models including bge-large-en-v1.5~\cite{bge_embedding}, and bge-reranker-large~\cite{bge_embedding}, as well as LLM-based ranking models such as RankLlama~\cite{ma2024rankllama}, RankVicuna~\cite{pradeep2023rankvicunazeroshotlistwisedocument}, RankZephyr~\cite{pradeep2023rankzephyreffectiverobustzeroshot}, FirstMistral~\cite{chen2024earlyreproductionimprovementssingletoken} and RankGPT (\texttt{gpt-4o}\footnote{\texttt{gpt-4o} refers to \texttt{gpt-4o-2024-08-06} from OpenAI})~\cite{sun2024chatgptgoodsearchinvestigating}. 

\noindent\textbf{Implementation Details}. {\Ours} is built upon Llama-3.1-8B-Instruct\footref{llama3.1-8b-inst}, trained for 5 epochs with an effective batch size of 512 and a learning rate of $5 \times 10^{-6}$ using AdamW optimizer~\cite{loshchilov2019decoupledweightdecayregularization}.
To evaluate the effectiveness of {\Ours}, we keep the first-stage retrieval and the generator fixed. The first-stage retrieval uses bge-large-en-v1.5~\cite{bge_embedding}, a high-performance retrieval model, while the generator is Llama-3.1-8B-Instruct\footnote{\url{https://huggingface.co/meta-llama/Llama-3.1-8B-Instruct}\label{llama3.1-8b-inst}} which we use without fine-tuning.
For RAG, we adopt the standard RAG framework~\cite{ram2023incontextretrievalaugmentedlanguagemodels} to generate answers based on the retrieved contexts.
All baselines are implemented utilizing the Rankify~\cite{abdallah2025rankify} toolkit\footnote{\url{https://github.com/DataScienceUIBK/Rankify}}.

\subsection{End-to-end QA Evaluation}
\label{sec:qa_eval}
The main results of various ranking models are presented in Table~\ref{tb:RAG}.
The results in \textsc{Retrieval Only} are derived solely from first-stage retrievers, while \textsc{Reranking} and \textsc{Set Selection} correspond to the results of re-ranking or selecting over the top-20 candidates retrieved by the respective first-stage retrievers.
Based on the results, the key observations are as follows:
(1) In terms of answer correctness, {\Ours} significantly outperforms all baselines by selecting the optimal set of passages from the retrieved passages, achieving notably higher F1 and Accuracy.
These results are quite impressive, considering that \Ours~uses 40-50\% fewer passages on average compared to baselines.
(2) The performance of LLM-based ranking baselines, such as RankLlama and RankZephyr, was less satisfactory. We observe that the relatively small bge-reranker-large~\cite{bge_embedding} performs comparably to, and in some cases even outperforms LLM-based baselines.
This is likely due to complex questions making retrieval more challenging, introducing more noise, and requiring not only relevance but also diversity, completeness, and comprehensiveness. 
We posit that the key to the RAG system lies in having more useful knowledge and fewer distracting passages, enabling even a simple and smaller model to outperform LLMs when the correct RAG paradigm is applied, which we will discuss in detail in (§\ref{subsec:set-selection}).

\begin{figure*}[ht]
\renewcommand{\arraystretch}{0.9}
  \begin{subfigure}{0.33\textwidth}
    \centering
    \includegraphics[width=.9\linewidth]{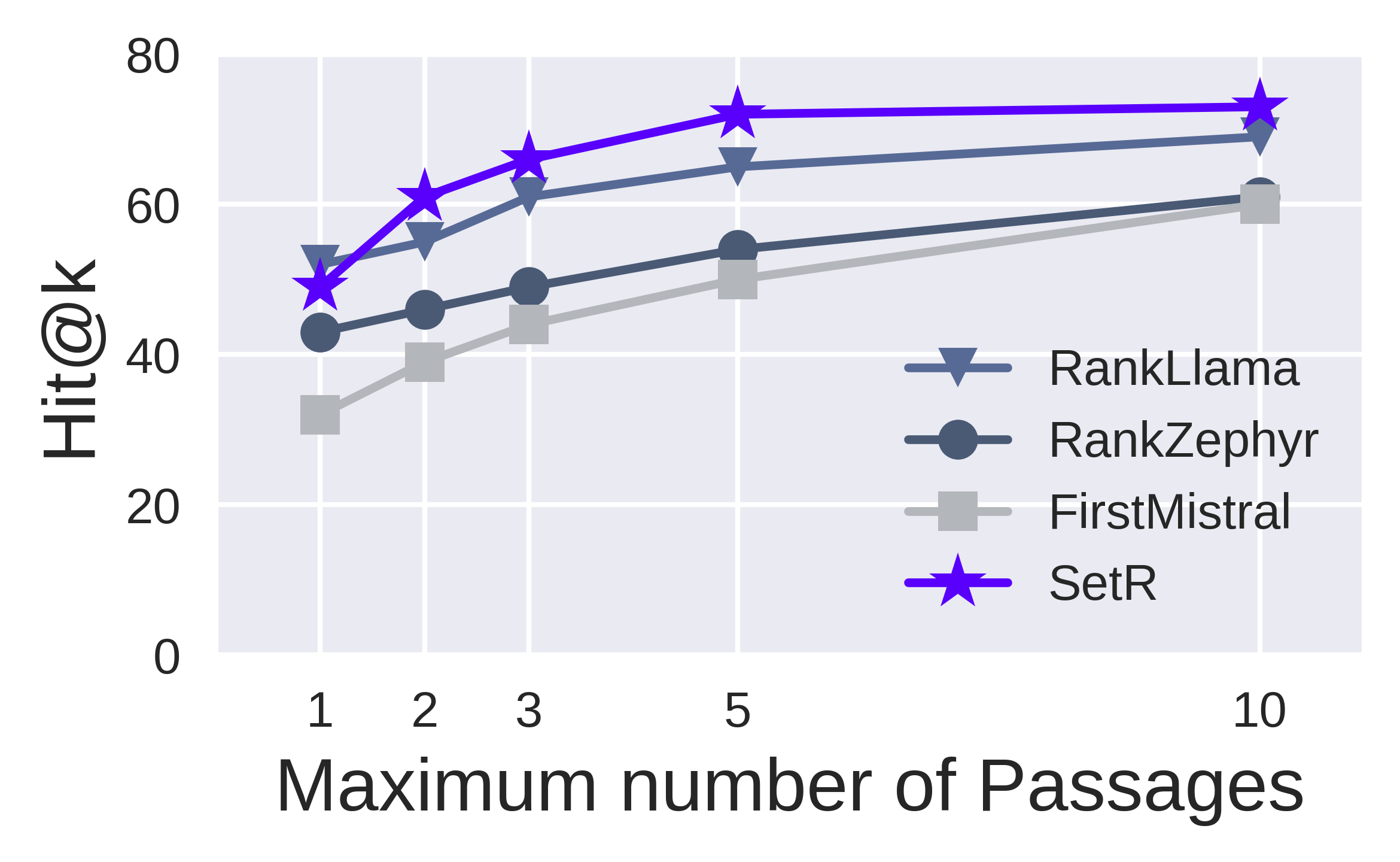}
    
    \caption{Hit@\textit{k} distribution}
    \label{fig:1}
  \end{subfigure}%
  \begin{subfigure}{0.33\textwidth}
    \centering
    \includegraphics[width=.9\linewidth]{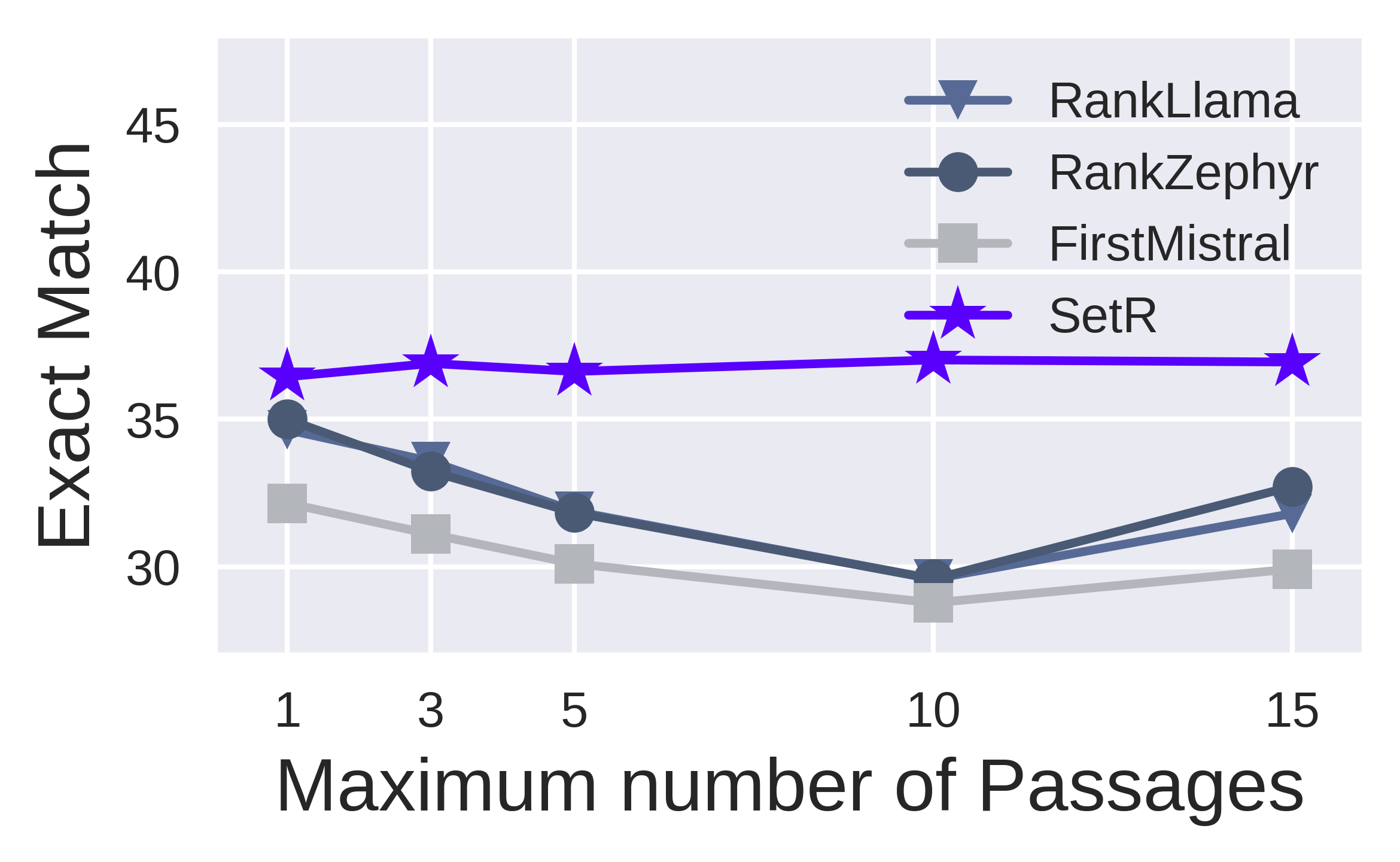}
    
    \caption{HotpotQA~\cite{yang2018hotpotqadatasetdiverseexplainable}}
    \label{fig:2}
  \end{subfigure}
  \begin{subfigure}{0.33\textwidth}\quad
    \centering
    \includegraphics[width=.9\linewidth]{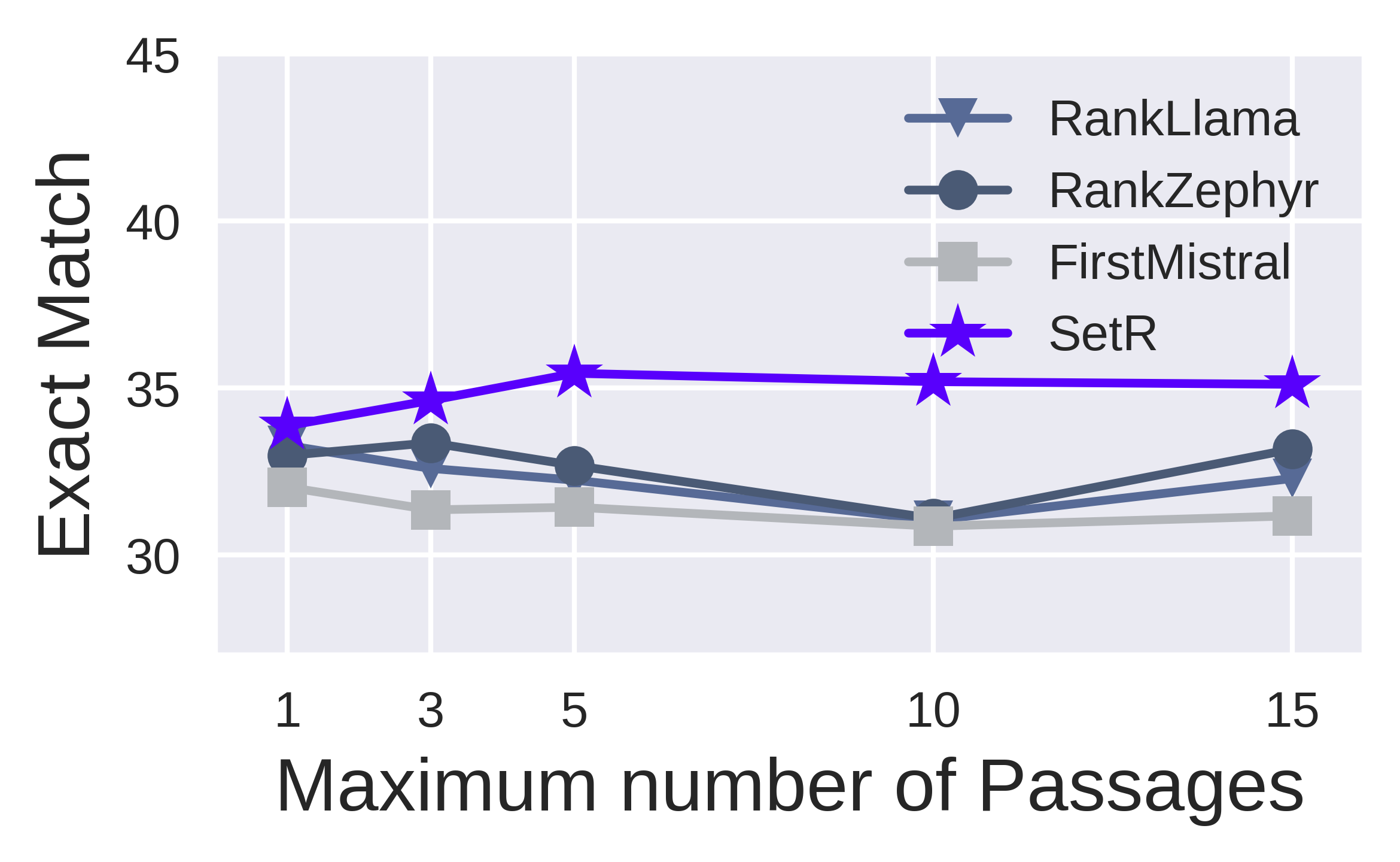}
    
    \caption{2WikiMultiHopQA~\cite{xanh2020_2wikimultihop}}
    \label{fig:3}
  \end{subfigure}
  \medskip

  \begin{subfigure}{0.33\textwidth}
    \centering
    \includegraphics[width=.9\linewidth]{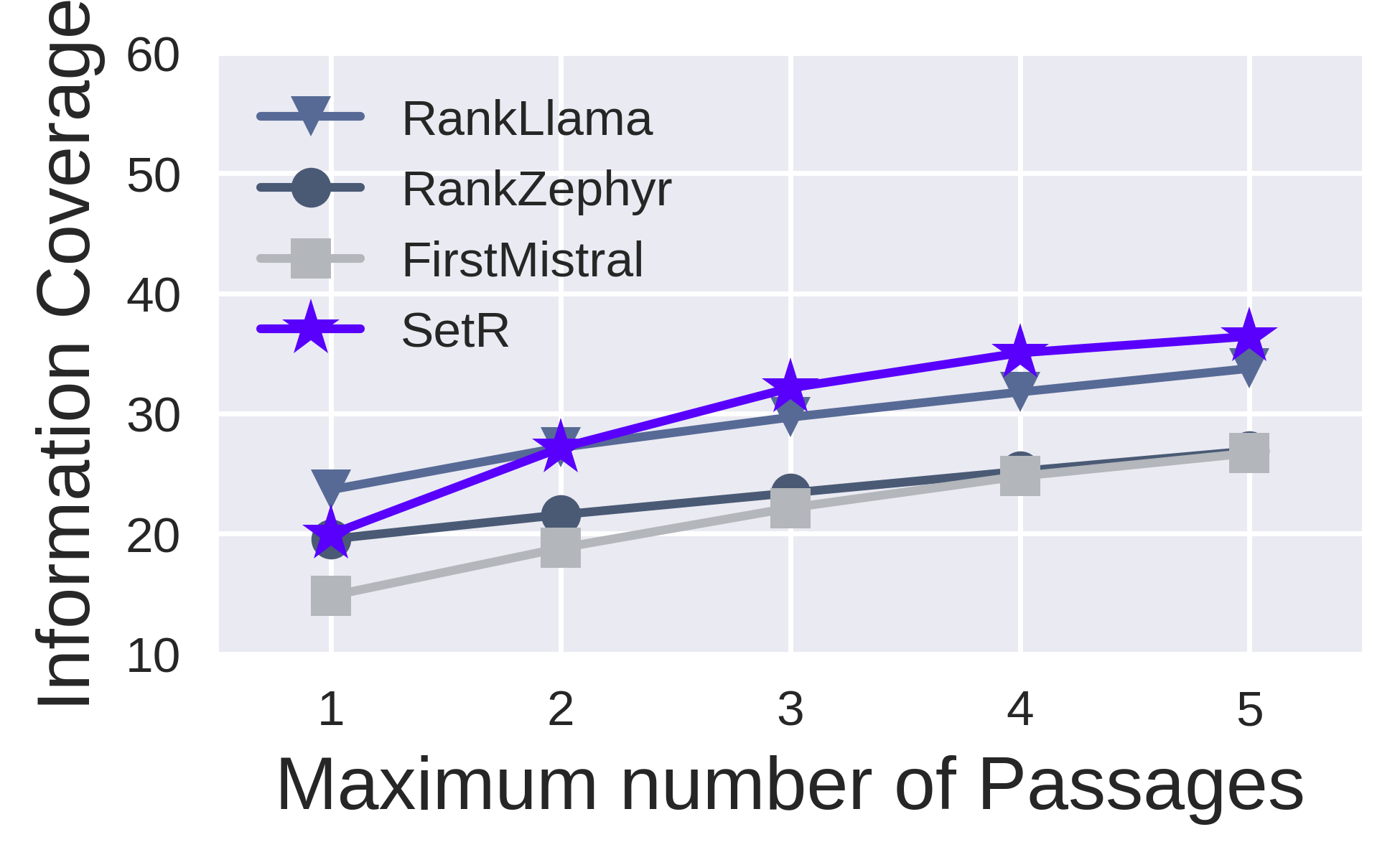}
    
    \caption{Average information coverage (\%)}
    \label{fig:4}
  \end{subfigure}
  \begin{subfigure}{0.33\textwidth}
    \centering
    \includegraphics[width=.9\linewidth]{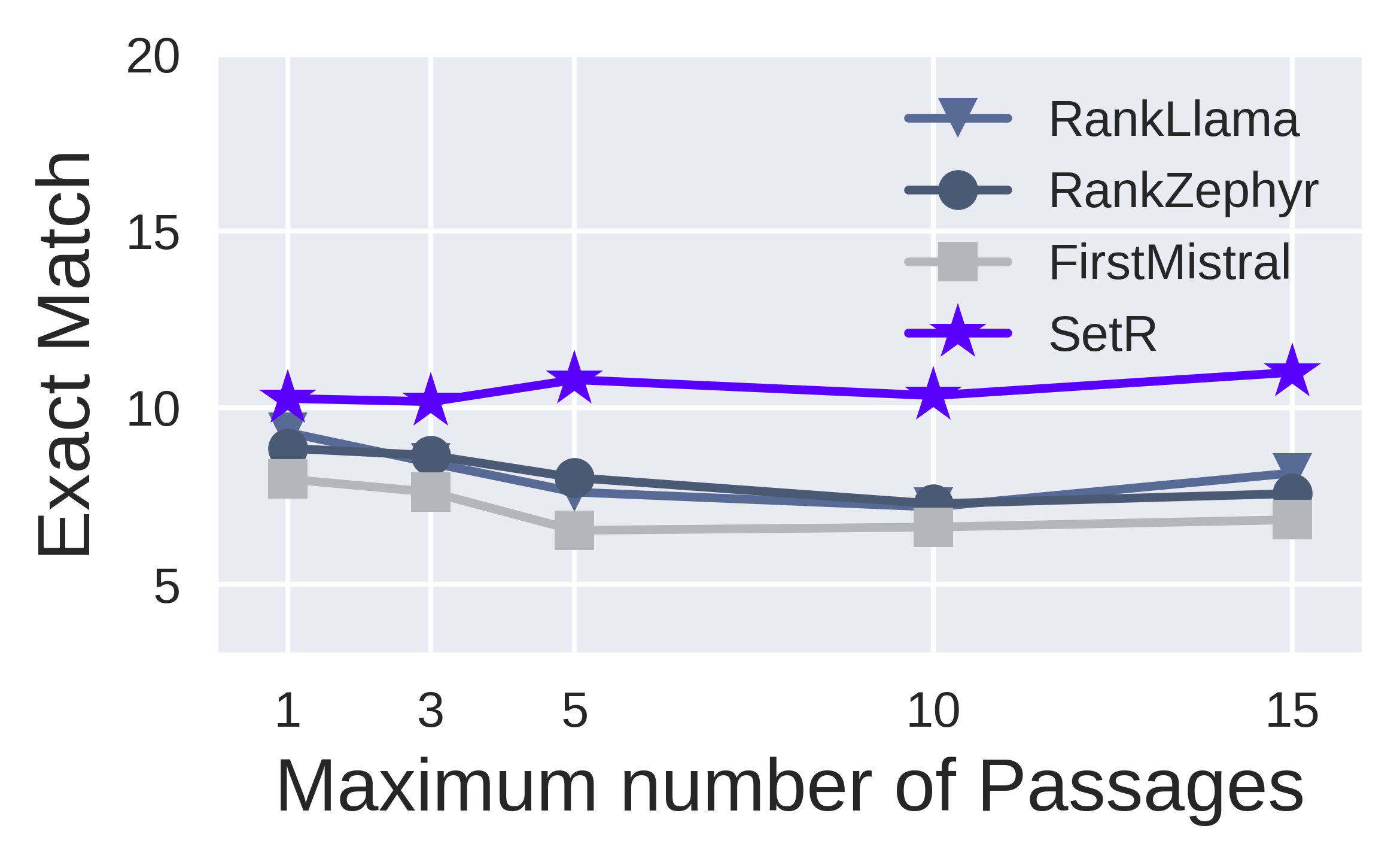}
    
    \caption{MusiQue~\cite{trivedi2021musique}}
    \label{fig:5}
  \end{subfigure}
  \begin{subfigure}{0.33\textwidth}
    \centering
    \includegraphics[width=.9\linewidth]{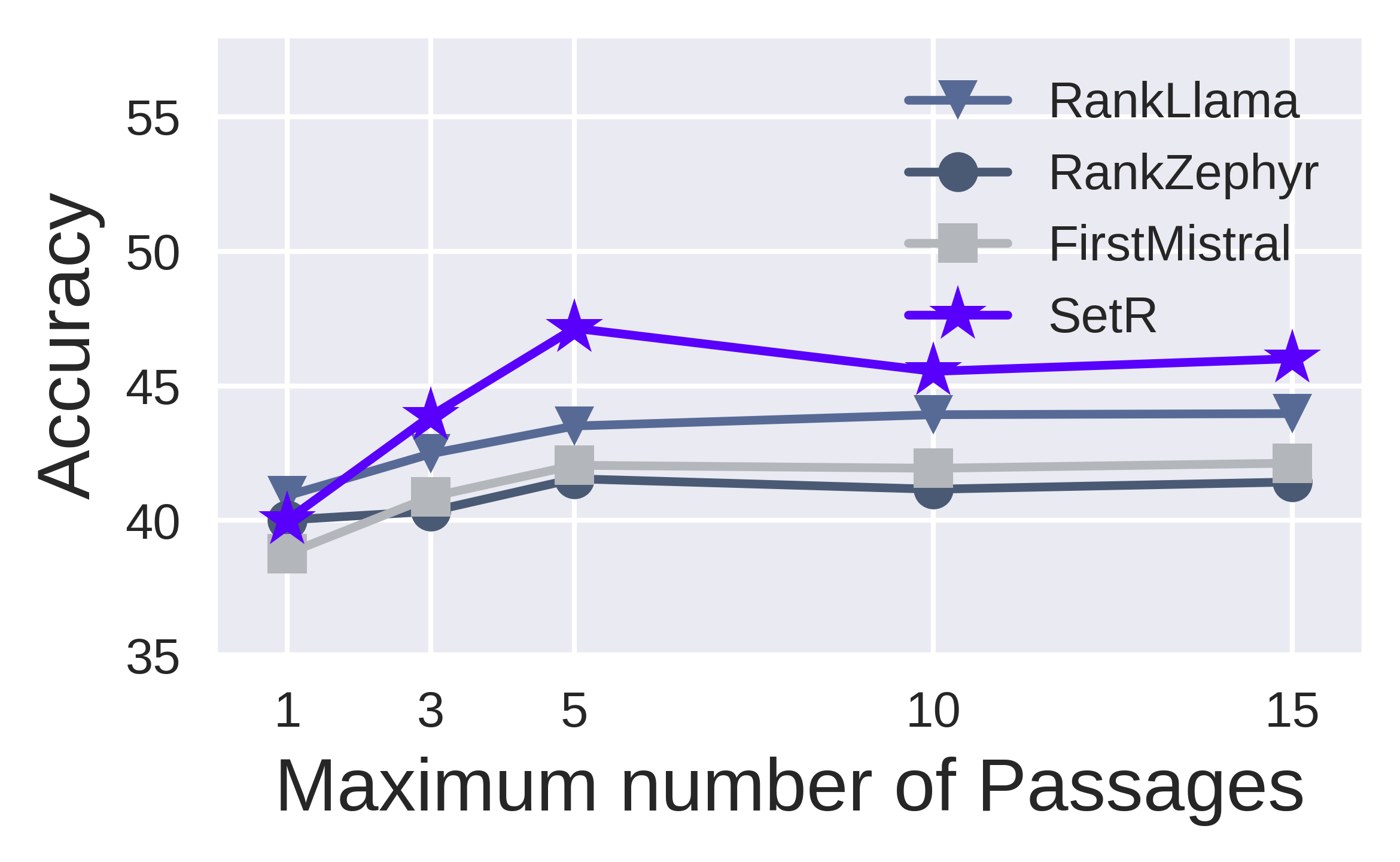}
    
    \caption{MultiHopRAG~\cite{tang2024multihoprag}}
    \label{fig:6}
  \end{subfigure}
  \caption{(a) Hit@\textit{k} distribution, and (d) average information coverage (\%) are measured based on gold evidence lists from the MultiHopRAG benchmark. (b), (c), (e), and (f) report results of the standard RAG pipeline across benchmarks, varying the maximum number of passages utilized for answer generation. Note that \Ours~utilizes fewer than 5 passages on average for answer generation.}
  \label{final_of_final}
\end{figure*}

\subsection{Retrieval Evaluation}
\label{sec:retrieval_eval}
The retrieval evaluation is conducted using both rank-based metrics such as MRR and NDCG, as well as presence-based selection metrics including Precision and Recall, on the MultiHopRAG dataset~\cite{tang2024multihoprag}.
The results of retrieval evaluation are shown in Table~\ref{tb:rerank}.
The results indicate that our model consistently achieves 3.8\%-4.6\% higher precision compared to off-the-shelf baselines, whereas it maintains competitive performance on rank-based metrics even with a small number of passages. 

LLM-based rankers, paired with advanced retrievers, perform well on rank-based metrics compared to the first-stage retrieval baselines.
Specifically, RankGPT (\texttt{gpt-4o}) shows notable improvements in both rank-based and presense-based metrics. 
However, we observe a discrepancy between rank-based metrics and end-to-end QA performance, as shown in Table~\ref{tb:RAG}.

\begin{table}[ht]
\resizebox{\linewidth}{!}{
\begin{tabular}{l|cccc}
\toprule
   \multirow{2}{*}{\textbf{Model}}
   & \multicolumn{2}{c}{\textsc{Rank-based}}  & \multicolumn{2}{c}{\textsc{Presence-Based}} 
   \\ 
   &MRR@10&NDCG@10& Prec@5 & Recall@5 \\
    \noalign{\hrule height 0.8pt} 

\rowcolor{lightgray!40}\multicolumn{5}{l}{\textsc{Retrieval Only}}\\
\hspace{3mm}BM25&0.4429&0.6827&0.1109&0.2413\\
\hspace{3mm}bge-large-en-v1.5&0.4523&0.6900&0.1612&0.3232\\
\hline
\rowcolor{lightgray!40}\multicolumn{5}{l}{\textsc{Reranking}}\\
\hspace{3mm}bge-reranker-large&0.6019&0.7481&0.1619&0.3276\\
\hspace{3mm}RankLlama&\underline{0.6311}&\textbf{0.7703}&0.1679&0.3375\\
\hspace{3mm}RankVicuna&0.5077&0.7232&0.1372&0.2760\\
\hspace{3mm}RankZephyr&0.5326&0.7046&0.1340&0.2685\\
\hspace{3mm}FirstMistral&0.4521&0.6895&0.1321&0.2671\\
\hspace{3mm}RankGPT (\texttt{gpt-4o})&\textbf{0.6358}&\underline{0.7628}&0.1799&\underline{0.3601}\\
\hline
\rowcolor{lightgray!40}\multicolumn{5}{l}
{\textsc{Set Selection (Ours)}} \\

\hspace{3mm}\Ours-Selection only&0.5610&0.7295&\underline{0.2187}&0.3554\\
\hspace{3mm}\Ours-CoT&0.5533&0.7281&0.2047&0.3413\\
\hspace{3mm}\Ours-CoT \& IRI&0.5742&0.7255&\textbf{0.2268}&\textbf{0.3669}\\

\bottomrule
\end{tabular}
}
\caption{
Retrieval performance on the MultiHopRAG benchmark. \textsc{Rank-based} metrics reflect passage order, while \textsc{Presence-Based} metrics consider only presence. Best and second-best scores are in \textbf{bold} and \underline{underlined}, respectively.
}
\label{tb:rerank}
\end{table}

This discrepancy arises from the assumption in rank-based metrics that relevance is determined on a one-to-one basis between a document and a query. 
For more complex question types, such as multi-hop questions, where considering relationships between multiple documents is needed for comprehensive information retrieval, these metrics may need further refinement to improve accuracy. 

\section{Analysis}
In this section, we analyze the key components that contribute to the effectiveness of our set-wise passage selection approach \Ours.
We evaluate the quality of selected passages in terms of information coverage and robustness (§\ref{subsec:set-selection}), assess the impact of reasoning strategies including information requirement identification (IRI) and Chain-of-Thought (CoT) §\ref{subsec:effectiveness}, design a controlled setup to isolate method-level effects (§\ref{subsec:equitable}), and compare token efficiency between selection-based and reranking-based methods (§\ref{subsec:efficiency}).



\subsection{Effectiveness of Set Selection: Informativeness and Robustness}
\label{subsec:set-selection}
We analyze the effectiveness of our selection approach in terms of two key perspectives: (1) informativeness; how comprehensively the selected passages cover the necessary information, and (2) robustness; the model's ability to discard irrelevant or redundant content.

\noindent \textbf{Informativeness}.
Traditional ranking-based retrieval systems typically score and select passages based on their individual relevance rather than the collective information gained from multiple passages, and often fail to account for content duplication. 
To address this, we additionally analyze \textit{information coverage}, measuring how \textit{newly} retrieved gold evidence accumulates as the number of selected documents increases.
This helps capture how \textit{valid} information accumulates as additional documents are included.
The MultiHopRAG dataset provides gold evidence lists, extractively collected from documents, allowing precise measurement of how much necessary information is covered by the selected passages.
The formula used to calculate information coverage is detailed in Appendix~\ref{ab:detail}.
We report two complementary metrics: Hit@\textit{k}, which reflects whether gold evidence appears in the top-k selected passages, and \textit{information coverage}, which measure the accumulation of distinct gold evidences across the selected evidences.



As illustrated in Figure~\ref{fig:1} and Figure~\ref{fig:4}, 
despite {\Ours} selects passages without any ordering, it outperforms reranking methods with a notable improvement at Hit@\textit{k} from 48.87\% to 69.90\%. 
Furthermore, \Ours~enhances information coverage from 19.33\% to 36.49\%, whereas other rerankers achieve only a modest average gain of 9.80\% in the acquisition of new information.
These findings suggest that reranking methods may struggle to retrieve documents containing previously uncovered information due to the presence of duplicate content and hard negatives. 
In contrast, our method captures both individual relevance and collective information coverage, which are essential for improving retrieval effectiveness and overall answer correctness.

\noindent \textbf{Robustness}.
We further assess the robustness of our approach to retrieval noise by analyzing how answer generation quality changes as the number of passages increases.
As shown in Figure~\ref{fig:2},~\ref{fig:3},~\ref{fig:5}, and ~\ref{fig:6}, simply increasing the number of input passages often leads to performance degradation, suggesting that more is not always better. 
This is especially pronounced in multi-hop QA, where irrelevant or contradictory evidence can mislead the generator.
In contrast, {\Ours} consistently achieves stronger performance while utilizing significantly fewer passages.
On average, it uses 2.91 passages compared to the standard top 5 used in reranking.
This demonstrates the model's ability to discriminate high-utility information from distractors, thereby maintaining both efficiency and factual precision.
These findings suggest that effective passage selection requires balancing recall and conciseness, and that a smaller, curated set can outperform longer, noisier contexts.

\subsection{Effectiveness of Reasoning Components: The Role of CoT and IRI}\label{subsec:effectiveness}






To understand the role of reasoning in passage selection, we conduct an ablation study with three \Ours~variants: (1) without any reasoning, (2) with general CoT reasoning, and (3) with our proposed IRI-based reasoning.

As presented in Table~\ref{tb:RAG} and Table~\ref{tb:rerank},
in terms of precision, we observe a significant performance improvement when applying IRI-based explicit requirement analysis and selection, compared to using standard CoT reasoning alone, or no reasoning process. 
Table~\ref{tb:ablation} shows a similar trend, applying IRI-based reasoning yields stronger end-to-end RAG performance compared to other methods; more details are in §\ref{subsec:equitable}.
%
These results suggest that the performance gains of our approach do not simply stem from leveraging the intrinsic thinking steps of LLMs, and highlight the critical role of our IRI step in assembling a passage set with maximum information coverage, demonstrating the best performance across all metrics and benchmarks.

\subsection{A More Equitable Comparison: Method-Level Effects}\label{subsec:equitable}
While prior sections demonstrate that our approach improves end-to-end performance, such comparison can still be influenced by external factors, including differences in base models, data sources, or teacher supervision.
To more rigorously assess the intrinsic effectiveness of our set-wise selection approach, we conduct a method-level evaluation that explicitly controls for these confounding factors.
Specifically, we implement two strategies to minimize the impact of confounding factors: (1) upper bound; a teacher model directly performs the selection task, to isolate the contribution of the method itself, 
and (2) unified setting; all baselines are retrained using the same base model, training data, and teacher supervision, ensuring a fair and method-focused comparison.


\begin{table}
\renewcommand{\arraystretch}{1.2}
\center

\resizebox{\columnwidth}{!}{
\centering
\begin{tabular}{l|cc|cc|cc|c}
\toprule
\multirow{2}{*}{\textbf{Method}}

&  \multicolumn{2}{c|}{\textbf{HotpotQA}} & \multicolumn{2}{c|}{\textbf{2Wiki}} & \multicolumn{2}{c|}{\textbf{MuSiQue}} & {\textbf{MHRAG}}  \\
&EM&F1&EM&F1&EM&F1&Accuracy\\ 
\hline
\multicolumn{8}{l}{\cellcolor{lightgray!40}\textsc{Reranking}} \\
\hspace{2mm}Rank only$^\blacklozenge$ & 33.85&34.45 & 34.36&28.06 & 9.43&13.25  & 45.69 \\
\hspace{2mm}Rank + CoT$^\blacklozenge$& 34.61&35.26 & 34.77&28.18 & 9.52&13.51  & 45.26 \\ 
\hline
\multicolumn{8}{l}
{\cellcolor{lightgray!40} {\textsc{Set Selection}}} \\
\hspace{2mm}\Ours-Selection only$^\blacklozenge$  & \underline{38.28}& \underline{40.12} & \underline{35.83}&\textbf{31.14} & \underline{11.50}&\underline{16.37} & \underline{46.24} \\
\hspace{2mm}\Ours-CoT$^\blacklozenge$    & 37.46&39.44 & \textbf{35.85}&31.07 & 10.79&15.56 & 44.95 \\
\hspace{2mm}\Ours-CoT \& IRI$^\blacklozenge$      & \textbf{39.16} &\textbf{40.49} & 35.68&\underline{31.09} & \textbf{12.33} &\textbf{ 16.91} & \textbf{46.40} \\ \bottomrule

\end{tabular}}

\caption{
Ablation study with teacher model inference to isolate method-level effects. All experiments use \texttt{gpt-4o-2024-08-06} as the reference model, annotated with $^\blacklozenge$. For \textsc{Reranking}, 
RankGPT4 prompt~\citep{sun2024chatgptgoodsearchinvestigating}
is employed. Prompts for \textsc{Set Selection} are provided in Appendix \ref{prompt-templates}. \textbf{2Wiki} and \textbf{MHRAG} refer to the 2WikiMultiHopQA and MultiHopRAG benchmarks, respectively.
}
\label{tb:ablation}
\end{table}

\noindent \textbf{Teacher model as Upper Bound.}\label{para:IRI}
We first design an upper-bound evaluation setup where all models, whether reranking or selection-based, leverage the same powerful teacher model, GPT-4o, to generate passage selections or ranking.
This setup isolates the effect of the selection formulation itself by minimizing the influence of model capacity or training-specific artifacts.
Table~\ref{tb:ablation}, which reports the method-level effects with teacher model, indicates that our set selection method outperforms traditional reranking strategies, highlighting the individual contributions of both CoT reasoning and IRI to enhanced retrieval performance.

Specifically, the \Ours~method with IRI consistently achieves strong performance across all benchmarks and demonstrates competitive or superior results compared to both traditional reranking and other set selection variants.
Notably, while both Rank + CoT and \Ours-CoT share the same CoT prompt, the results reveal that using set-wise selection rather than integrated ranking in the final stage leads to improved retrieval outcomes.
This suggests that constructing a unified ranking may lead to the omission of certain aspects of the reasoning process, potentially resulting in information loss.
%
Furthermore, the comparison between \Ours-CoT and \Ours-CoT \& IRI demonstrates that explicitly identifying essential information during the reasoning process helps improve selection precision and ensures broader information coverage.

\begin{table}
\renewcommand{\arraystretch}{1.2}
\center
\resizebox{\columnwidth}{!}{
\centering
\begin{tabular}{l|cc|cc|cc|c}
\toprule
\multirow{2}{*}{\textbf{Model}}
&  \multicolumn{2}{c|}{\textbf{HotpotQA}} & \multicolumn{2}{c|}{\textbf{2Wiki}} & \multicolumn{2}{c|}{\textbf{MuSiQue}} & {\textbf{MHRAG}}  \\
&EM&F1&EM&F1&EM&F1&Accuracy \\
\hline
\multicolumn{8}{l}{\cellcolor{lightgray!40}{\textsc{Built on Zephyr-7B-$\beta$}}~\citep{pradeep2023rankzephyreffectiverobustzeroshot}}\\
\hspace{2mm}RankZephyr (original) & 29.76 & 30.36 &31.19&24.92&6.95&10.57&41.55 \\
\hline
\multicolumn{8}{l}
{\cellcolor{lightgray!40}{\textsc{Built on Llama-3.1-8B-Instruct}}} \\
\hspace{2mm}RankZephyr$^\clubsuit$ &34.69&35.04&33.87&27.83&8.61&12.79&43.90 \\
\hspace{2mm}RankZephyr + CoT$^\clubsuit$ &33.99&34.38&33.66&27.85&9.43&13.27&43.60\\

\hspace{2mm}\Ours-CoT \& IRI & \textbf{36.62}&\textbf{38.11}&\textbf{35.44}&\textbf{30.35}&\textbf{10.79}&\textbf{15.43}&\textbf{47.14}\\
\bottomrule
\end{tabular}
}

\caption{Fair comparisons under a unified setting, with confounding factors minimized.
RankZephyr$^\clubsuit$ and RankZephyr + CoT$^\clubsuit$ were implemented using the same LLaMA-3.1-8B-Instruct model as our method, fine-tuned on data re-annotated by \texttt{gpt-4o-2024-08-06}.
}
\label{tb:fair}
\end{table}


\begin{table*}[ht]
\resizebox{\linewidth}{!}{
\begin{tabular}{l|cc|cc|cc|cc}
\toprule
   \multirow{3}{*}{\textbf{Method}}
   & \multicolumn{2}{c}{\textbf{MultiHopRAG}}  & \multicolumn{2}{c}{\textbf{HotpotQA}} & \multicolumn{2}{c}{\textbf{2WikiMultiHopQA}} & \multicolumn{2}{c}{\textbf{MuSiQue}}
   \\ 
   &Reranker &Generator  &Reranker &Generator  &Reranker &Generator  &Reranker &Generator  \\
   &Output & Input&Output & Input&Output & Input&Output & Input\\
    \noalign{\hrule height 0.8pt} 

\rowcolor{lightgray!40}\multicolumn{9}{l}{\textsc{Direct inference}}\\
\hspace{3mm}RankZephyr (Llama-3.1-8B-Inst) &80 & 2672 & 80 & 1426 & 80 & 1504 & 80 & 1403\\
\hspace{3mm}\Ours-Selection only&17&1441&11&461&10&422&11&499\\
\hline
\rowcolor{lightgray!40}\multicolumn{9}{l}{\textsc{With reasoning}}\\
\hspace{3mm}RankZephyr + CoT (Llama-3.1-8B-Inst)& 603 & 2665 & 574 & 1435 & 560 & 1434 &599 & 1385\\
\hspace{3mm}\Ours-CoT&396&1517&383&426&376&333&412&439\\
\hspace{3mm}\Ours-CoT \& IRI&409&1240&317&432&276&332&340&503\\

\bottomrule
\end{tabular}
}
\caption{Efficiency analysis on token usage. We report the number of output tokens generated during retrieval and the number of input tokens fed into the generator, both of which serve as proxies for computational efficiency.}
\label{tb:efficiency}
\end{table*}

\noindent \textbf{Unified Setting for Direct Comparison.}
To further minimize potential bias, we create a unified training setting, in which all models are (1) built on the same base architecture (Llama-3.1-8B-Instruct), (2) trained on the same teacher supervision (re-annotated with \texttt{gpt-4o-2024-08-06}), and (3) evaluated using identical generation protocols. 
As shown in Table~\ref{tb:fair}, \Ours~again surpasses reranking baselines in end-to-end QA accuracy across all tasks, reaffirming the advantages of selection-based approaches.
%
%
This carefully controlled comparison highlights a critical insight: ranking-based reasoning, even with CoT, tends to collapse multiple reasoning chains into a single score, potentially obscuring key informational elements.
In contrast, our selection strategy preserves intermediate reasoning steps, resulting in better alignment with the actual information needs of the question.
These findings collectively demonstrate that the observed performance gains stem not from stronger base models or teacher supervision alone, but from our methodological reformulation of retrieval as set selection grounded in reasoning.

\subsection{Efficiency Analysis}\label{subsec:efficiency}
An ideal retrieval method for RAG systems should not only achieve strong performance—yielding accurate results—but also be efficient, minimizing latency, memory usage, and compute cost for real-world deployment.
While direct measurement of GPU time or inference latency depends on hardware and implementation choices, token-level analysis offers a practical proxy for computational efficiency.
We compare token usage across models by measuring: (1) the number of input tokens passed to the generator, and (2) the number of output tokens generated during the retrieval stage.
As shown in Table~\ref{tb:efficiency}, all \Ours~variants require substantially fewer input tokens than reranking-based methods.
For instance, on MultiHopRAG, \Ours-CoT \& IRI feeds only 1,240 input tokens into the generator, compared to 2,672 in RankZephyr.
Despite this sharp reduction, {\Ours} not only maintains answer quality but often surpasses reranking models in F1 and accuracy.
Interestingly, even within {\Ours} variants, reasoning plays a role in efficiency. \Ours-Selection only is the most efficient in terms of token usage, while \Ours-CoT \& IRI trades some marginal increase in prompt length for greater answer correctness and recall.
This suggests a useful accuracy-efficiency trade-off spectrum that practitioners can tune based on resource constraints and latency budgets.

\section{Discussion}

Our work highlights the limitations of conventional top-$k$ retrieval in Retrieval-Augmented Generation (RAG) systems and proposes a set-wise passage selection approach to better address the unique information needs of generative models. By incorporating information requirement identification and Chain-of-Thought reasoning for passage selection, {\Ours}~(\underline{Set}-wise passage selection for \underline{R}etrieval-Augmented Generation), enhances retrieval precision and improves end-to-end answer accuracy.

A promising direction for future work is enabling efficient selection from large candidate passage pools (e.g., 100 passages). Unlike traditional listwise reranking, which often relies on sequential sliding windows, our approach supports parallel, order-agnostic selection—reducing time complexity and improving scalability.
Additionally, future work could explore how {\Ours} iteratively refines queries for better retrieval and develops adaptive selection techniques to adjust the number of selected passages based on domain-specific needs.



\section*{Limitations}


While our set-wise passage selection approach improves retrieval quality and efficiency in RAG systems, it has several limitations.

First, our method relies on a predefined retrieval pipeline, meaning its performance is still dependent on the initial retrieval stage. If the initial retrieved set lacks critical information, even an optimal set-wise selection cannot compensate for missing knowledge. 

Second, our method optimizes retrieval for multi-hop and complex queries but has not yet been validated across diverse RAG domains such as code generation or conversational AI. While we anticipate broader applicability, further evaluation on diverse tasks is needed.

Lastly, our approach relies on the quality of the underlying language model for reasoning-based selection. While chain-of-thought reasoning improves passage selection, its effectiveness depends on the LLM’s ability to accurately analyze and synthesize information. 

\bibliography{custom}
\appendix
\section{Appendix}
\label{sec:appendix}
\subsection{Datasets}
The experiments were conducted on the following four benchmark datasets:

\begin{itemize}
  \item \textbf{HotpotQA}~\cite{yang2018hotpotqadatasetdiverseexplainable} is a large-scale multi-hop QA dataset with 113k question–answer pairs from Wikipedia. Each question requires reasoning over multiple documents, with sentence-level supporting facts provided. It includes diverse queries and comparison questions that test compositional reasoning and explainability, making it a strong benchmark for multi-hop retrieval systems.
  \item \textbf{2WikiMultiHopQA}~\cite{xanh2020_2wikimultihop}  is a multi-hop QA dataset combining Wikipedia text with Wikidata triples to evaluate step-by-step reasoning. Each question includes an explicit reasoning path linking entities across documents. The dataset tests compositional reasoning and requires models to use both unstructured text and structured knowledge to answer questions.
  \item \textbf{MusiQue}~\cite{trivedi2021musique} is a multi-hop QA benchmark designed to prevent reasoning shortcuts by requiring genuine multi-step reasoning. It includes around 25k questions composed from connected single-hop queries, each requiring 2 to 4 reasoning steps. The dataset emphasizes strong logical dependencies between steps and includes unanswerable variants to test robustness. MuSiQue is more challenging than earlier datasets and highlights significant performance gaps between humans and models.
  \item \textbf{MultiHopRAG}~\cite{tang2024multihoprag} is a recent benchmark for evaluating retrieval-augmented generation on complex multi-hop queries. Unlike the above QA datasets, MultiHopRAG is specifically built to evaluate end-to-end RAG systems. It includes 2,556 questions over a corpus of English news articles, with answers supported by evidence from 2 to 4 documents. The queries involve temporal and entity-based reasoning and require retrieving and synthesizing information across multiple sources. This makes the dataset well-suited for evaluating our set-wise passage selection approach.
\end{itemize}

\subsection{Baselines}
Following reranking models are considered as baselines:
\begin{itemize}
\item \textbf{bge-reranker-large}~\cite{bge_embedding} is a lightweight cross-encoder model from BAAI that scores query–passage pairs using full cross-attention, enabling more accurate relevance judgments than embedding-based models. Fine-tuned on large-scale data, it is commonly used to rescore top-$k$ results in retrieval pipelines. 
As a strong open-source baseline, it reflects state-of-the-art conventional reranking focused on individual passage relevance.
  \item \textbf{RankLlama}~\cite{ma2024rankllama}  is a pointwise reranker based on LLaMA-2 7B model~\cite{touvron2023llama2openfoundation}.
Given a query and a candidate passage, it outputs a score to reorder retrieved documents by their relevance.
It demonstrates strong performance in both in-domain and zero-shot settings, serving as a competitive open-source baselines for passage reranking.
  \item \textbf{RankVicuna}~\cite{pradeep2023rankvicunazeroshotlistwisedocument} is a 7B open-source listwise reranker built on the Vicuna 7B model~\cite{zheng2023judgingllmasajudgemtbenchchatbot}. 
  It takes a query and a list of passages as input and outputs a ranked list of passage indices.
  Trained with GPT generated supervision, it achieves performance comparable to GPT-3.5 on benchmarks like TREC DL ~\cite{craswell2020overviewtrec2019deep,craswell2021overviewtrec2020deep}, providing a transparent alternative to proprietary rerankers.
  \item \textbf{RankZephyr}~\cite{pradeep2023rankzephyreffectiverobustzeroshot} is a zero-shot listwise reranker built on the Zephyr-7B model~\cite{tunstall2023zephyrdirectdistillationlm}.
  Fine-tuned using GPT-4 generated ranking, it outputs ordered lists of passage indices given a query and candidate passages.
  It achieves performance close to GPT-4 and even surpassing it on some benchmarks.
  Its open-source nature and reproducibility make it a robust baseline for evaluating listwise reranking methods.
  \item \textbf{FirstMistral}~\cite{chen2024earlyreproductionimprovementssingletoken} is a zero-shot listwise reranker based on Mistral 7B~\cite{jiang2023mistral7b}. It reframes reranking as a single-token decoding task, enabling fast and efficient passage selection. Despite its simplicity, it achieves competitive performance and serves as a strong open-source baseline to assess the raw ranking ability of modern instruction-tuned LLMs.
  \item \textbf{RankGPT4}~\cite{sun2024chatgptgoodsearchinvestigating} is a GPT-4-based reranker accessed via OpenAI’s API, used in a zero-shot setting to rank passages given a query. It delivers state-of-the-art performance but is closed-source, non-reproducible, and costly. RankGPT4 serves as an upper-bound baseline to evaluate how well our approach performs against the strongest proprietary reranker.
\end{itemize}
\subsection{Additional Experimental Details}\label{ab:detail}
\noindent \textbf{Training}.
All \Ours~variants are fine-tuned using Llama-3.1-8B-Instruct as the base model. 
Training is conducted for 5 epochs using AdamW optimizer with a learning rate of $5 \times 10^{-6}$ and an effective batch size of 512. 
We use 16$\times$A100 GPUs and utilize Axolotl\footnote{\url{https://github.com/axolotl-ai-cloud/axolotl}} framework, which integrates various efficiency-oriented training techniques.
Each model is trained on 40k GPT-4o generated examples, where each input prompt includes a query and 20 retrieved passages.


\noindent \textbf{Evaluation Metrics}. We evaluate model performance using both retrieval and QA metrics. 
For retrieval, we report Mean Receprocal Rank (MRR), Normalized Discounted Cumulative Gain (NDCG), Precision, and Recall, which collectively measure the quality of passage ranking and coverage of relevant information.
For end-to-end QA, we adopt Exact Match (EM), F1 score, and Accuracy to quantify answer correctness and completeness.

Additionally, we evaluate \textit{information coverage} ($\textit{IC}@k$) on MultiHopRAG benchmark as it provides gold evidence lists, extractively collected from documents.

{\small\begin{alignat*}{1}
\textit{IC}@k = \frac{|\bigcup_{i=1}^{k}\{ e | e \in p_{i}\} \cap \{ e | e \in \mathcal{E}_{\text{gold}}\}|}{|\mathcal{E}_{\text{gold}}|}
\end{alignat*}}

where $p_i$ is the $i$-th top-ranked passage, $e$ is an evidence span, $\mathcal{E}_{\text{gold}}$ is the complete set of annotated gold evidences.

For each question, we collect the complete set of gold evidence $\mathcal{E}_{\text{gold}}$ required to answer it, as annotated in the MultiHopRAG dataset. Given a top-$k$ set of retrieved passages, we identify gold evidence spans $e$ within the passage text using regular expression matching and remove duplicated evidences if redundant passage exist. 
$\textit{IC}@k$ is then computed as the proportion of gold evidence that appears within the top-$k$ passages.

\begin{figure}[!th]
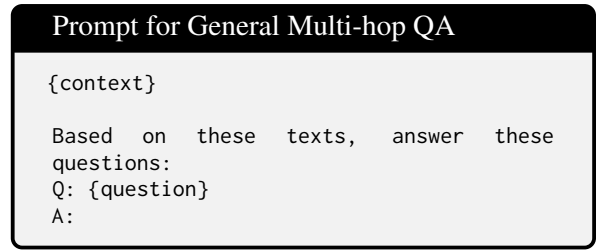

\begin{tcolorbox}[
    colback=gray!10,
    colframe=black,
    fontupper=\ttfamily\footnotesize,
    title={Prompt for General Multi-hop QA},
]
\{context\}
\\
\\
Based on these texts, answer these questions: 
\\
Q: \{question\}
\\
A:
\end{tcolorbox}
\vspace{-1em}
\caption{Prompt template used for general multi-hop QA datasets including HotpotQA, 2WikiMultiHopQA, and MusiQue.}
\label{fig:generation_general}
\end{figure}

\noindent \textbf{Generation}.
For answer generation, we utilize Rankify~\cite{abdallah2025rankify} toolkit to implement the full RAG pipeline, which consists of three components: Retrieval, Reranking/Selection, and Generation. 
For retrieval, we use the bge-large-en-v1.5~\cite{bge_embedding} model to retrieve the top-20 passages per query. 
These candidates are then reranked or filtered via a set-wise selection mechanism.
For generation, we employ Llama-3.1-8B-Instruct to generate final answers on general multi-hop QA benchmarks, including HotpotQA, 2WikiMultiHopQA, and MusiQUE.
For MultiHopRAG benchmark, we follow the evaluation protocol of~\cite{tang2024multihoprag} and employ \texttt{gpt-4o-2024-08-06} as the generator due to its strong reasoning performance. 
Notably, even when provided with gold evidences, open-source models such as Llama2-70B and Mixtral-8x7B achieve relatively low accuracy (0.32 and 0.36, respectively), while GPT-4 attains a significantly higher score of 0.89, highlighting a substantial performance gap.
The prompts used for each benchmark are shown in Figure~\ref{fig:generation_general} and Figure~\ref{fig:generation_multihop}, respectively.

\begin{figure}[!th]
\begin{tcolorbox}[
    colback=gray!10,
    colframe=black,
    fontupper=\ttfamily\footnotesize,
    title={Prompt for MultiHopRAG},
]
Below is a question followed by some context from different sources. Please answer the question based on the context. The answer to the question is a word or entity. If the provided information is insufficient to answer the question, respond `Insufficient Information'. Answer directly without explanation.
\\
\\
Question: \{question\}
\\
\\
Context: 
\\
\\
\{context\}
\end{tcolorbox}
\vspace{-1em}
\caption{Prompt template used for MultiHopRAG.}
\label{fig:generation_multihop}
\end{figure}

\subsection{Prompt Templates}
\label{prompt-templates}
We provide the prompt template used in our experiment for both \Ours-CoT and \Ours-Selection only.
Each prompt consists of a question and a set of passages, where each passage is assigned a unique numerical identifier.
In \Ours-CoT, the prompt concludes with a CoT reasoning, ``Let's think step by step.'', to encourage intermediate reasoning before producing the final selection.
In contrast, \Ours-Selection only removes this reasoning step and directly instructs the model to output only the final selection without any explanation.
Figure~\ref{fig:app_setr_cot} and~\ref{fig:app_selection_only} illustrate the two prompts, respectively.

\label{sec:appendix_implementation_details}

\begin{figure}[!th]
\begin{tcolorbox}[
    colback=gray!10,
    colframe=black,
    fontupper=\ttfamily\footnotesize,
    title={Prompt for {\Ours}-CoT},
]

I will provide you with \{num\} passages, each indicated by a numerical identifier []. Select the passages based on their relevance to the search query: \{question\}.\\
\\
\{context\}
\\
\\
Search Query: \{question\}
\\
\\
Select the passages that mostly cover clear and diverse information to answer the query. Number of passages is unlimited. 
\\The format of final output should be `\#\#\# Final Selection: [] []', e.g., \#\#\# Final Selection: [2] [1].\\
Let's think step by step.
\end{tcolorbox}
\vspace{-1em}
\caption{The set-wise passage selection prompt with basic Chain-of-Thought for \Ours-CoT.}
\label{fig:app_setr_cot}
\end{figure}

\begin{figure}[!th]
\begin{tcolorbox}[
    colback=gray!10,
    colframe=black,
    fontupper=\ttfamily\footnotesize,
    title={Prompt for {\Ours}-Selection only},
]

I will provide you with \{num\} passages, each indicated by a numerical identifier []. Select the passages based on their relevance to the search query: \{question\}.\\
\\
\{context\}
\\
\\
Search Query: \{question\}
\\
\\
Select the passages that mostly cover clear and diverse information to answer the query. Number of passages is unlimited. The format of final output should be `\#\#\# Final Selection: [] []', e.g., \#\#\# Final Selection: [2] [1]. \\
Only respond with the selection results, do not say any word or explain.
\end{tcolorbox}
\vspace{-1em}
\caption{The set-wise passage selection prompt without CoT process for \Ours-Selection only.}
\label{fig:app_selection_only}
\end{figure}






\end{document}